\documentclass[conference]{IEEEtran}
\IEEEoverridecommandlockouts

\usepackage{cite}
\usepackage{amsmath,amssymb,amsfonts}
\usepackage{algorithmic}
\usepackage{graphicx}
\usepackage{textcomp}
\usepackage{xcolor}
\usepackage{tabularray}
\usepackage{caption}
\usepackage{hyperref}
\usepackage[skip=-0in]{subcaption}
\def\BibTeX{{\rm B\kern-.05em{\sc i\kern-.025em b}\kern-.08em
    T\kern-.1667em\lower.7ex\hbox{E}\kern-.125emX}}

\begin{document}

\title{Reducing the Sensitivity of Neural Physics Simulators to Mesh Topology via Pretraining\\

\thanks{
Distribution Statement A. Approved for public release. Distribution is unlimited. This material is based upon work supported by the Under Secretary of Defense for Research and Engineering under Air Force Contract No. FA8702-15-D-0001. Any opinions, findings, conclusions or recommendations expressed in this material are those of the author(s) and do not necessarily reflect the views of the Under Secretary of Defense for Research and Engineering. © 2024 Massachusetts Institute of Technology. Delivered to the U.S. Government with Unlimited Rights, as defined in DFARS Part 252.227-7013 or 7014 (Feb 2014). Notwithstanding any copyright notice, U.S. Government rights in this work are defined by DFARS 252.227-7013 or DFARS 252.227-7014 as detailed above. Use of this work other than as specifically authorized by the U.S. Government may violate any copyrights that exist in this work.

© 2025 IEEE.  Personal use of this material is permitted. Permission from IEEE must be obtained for all other uses, in any current or future media, including reprinting/republishing this material for advertising or promotional purposes, creating new collective works, for resale or redistribution to servers or lists, or reuse of any copyrighted component of this work in other works.
}
}

\author{
\IEEEauthorblockN{Nathan Vaska}
\IEEEauthorblockA{
\textit{MIT Lincoln Laboratory}\\
Lexington, USA \\
nathan.vaska@ll.mit.edu}
\and
\IEEEauthorblockN{Justin Goodwin}
\IEEEauthorblockA{
\textit{MIT Lincoln Laboratory}\\
Lexington, USA \\
jgoodwin@ll.mit.edu}
\and
\IEEEauthorblockN{Robin Walters}
\IEEEauthorblockA{
\textit{Northeastern University}\\
Boston, USA \\
r.walters@northeastern.edu}
\and
\IEEEauthorblockN{Rajmonda S. Caceres}
\IEEEauthorblockA{
\textit{MIT Lincoln Laboratory}\\
Lexington, USA \\
rajmonda.caceres@ll.mit.edu}
}

\maketitle

\begin{abstract}

Meshes are used to represent complex objects in high fidelity physics simulators across a variety of domains, such as radar sensing and aerodynamics. There is growing interest in using neural networks to accelerate physics simulations, and also a growing body of work on applying neural networks directly to irregular mesh data. Since multiple mesh topologies can represent the same object, mesh augmentation is typically required to handle topological variation when training neural networks. Due to the sensitivity of physics simulators to small changes in mesh shape, it is challenging to use these augmentations when training neural network-based physics simulators. In this work, we show that variations in mesh topology can significantly reduce the performance of neural network simulators. We evaluate whether pretraining can be used to address this issue, and find that employing an established autoencoder pretraining technique with graph embedding models reduces the sensitivity of neural network simulators to variations in mesh topology. Finally, we highlight future research directions that may further reduce neural simulator sensitivity to mesh topology.

\end{abstract}


\section{Introduction}

\begin{figure}[ht]
\centering
    \hspace{.2in}
    \subfloat[Simulation Object]{%
        \begin{minipage}[c][1.43in]{0.2\textwidth}
        \centering
        \includegraphics[width=\textwidth]{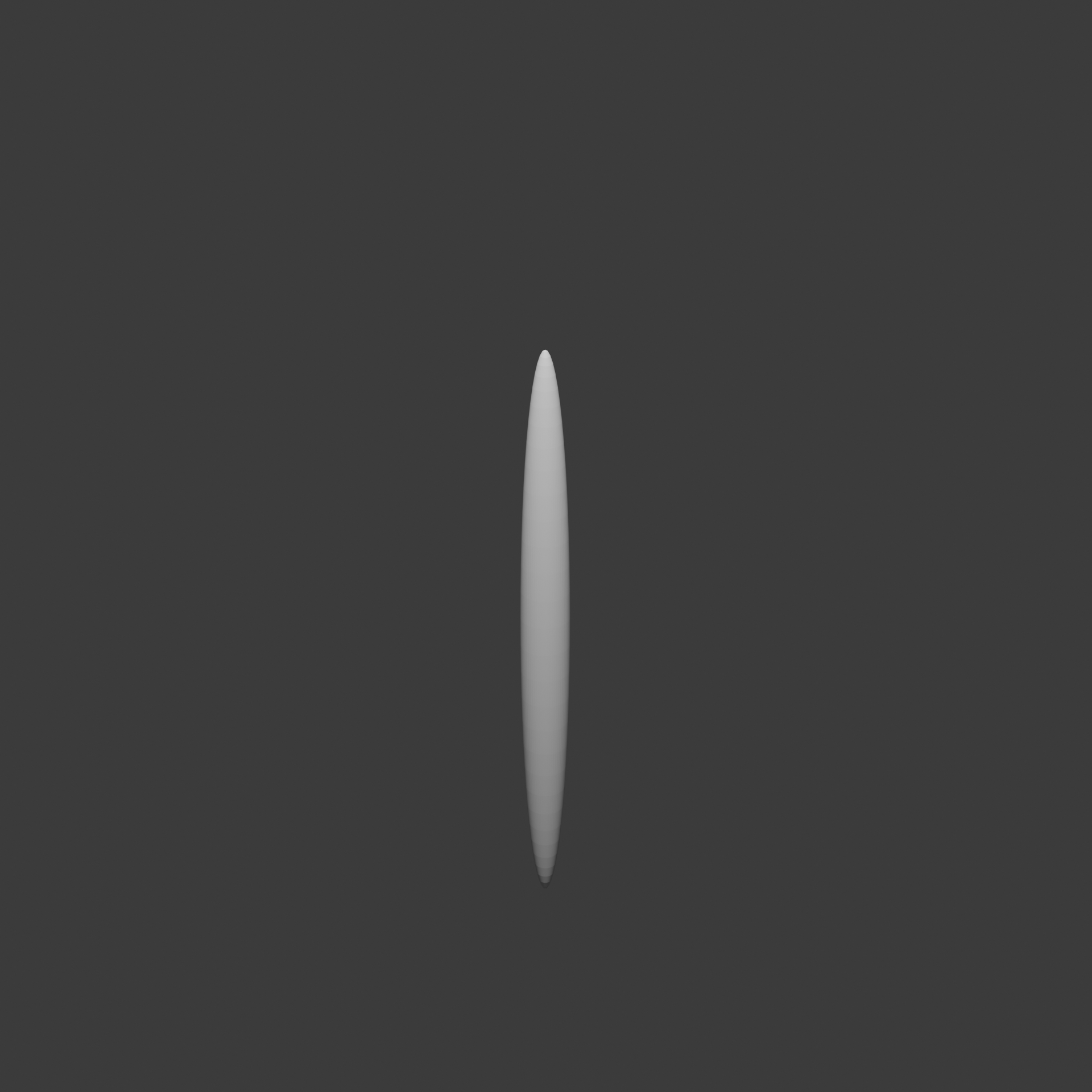}
        \label{aug:fig:a}
    \end{minipage}}
    \hspace{0.1in}
    \subfloat[Augmented Object]{%
        \begin{minipage}[c][1.43in]{0.2\textwidth}
        \centering
        \includegraphics[width=\textwidth]{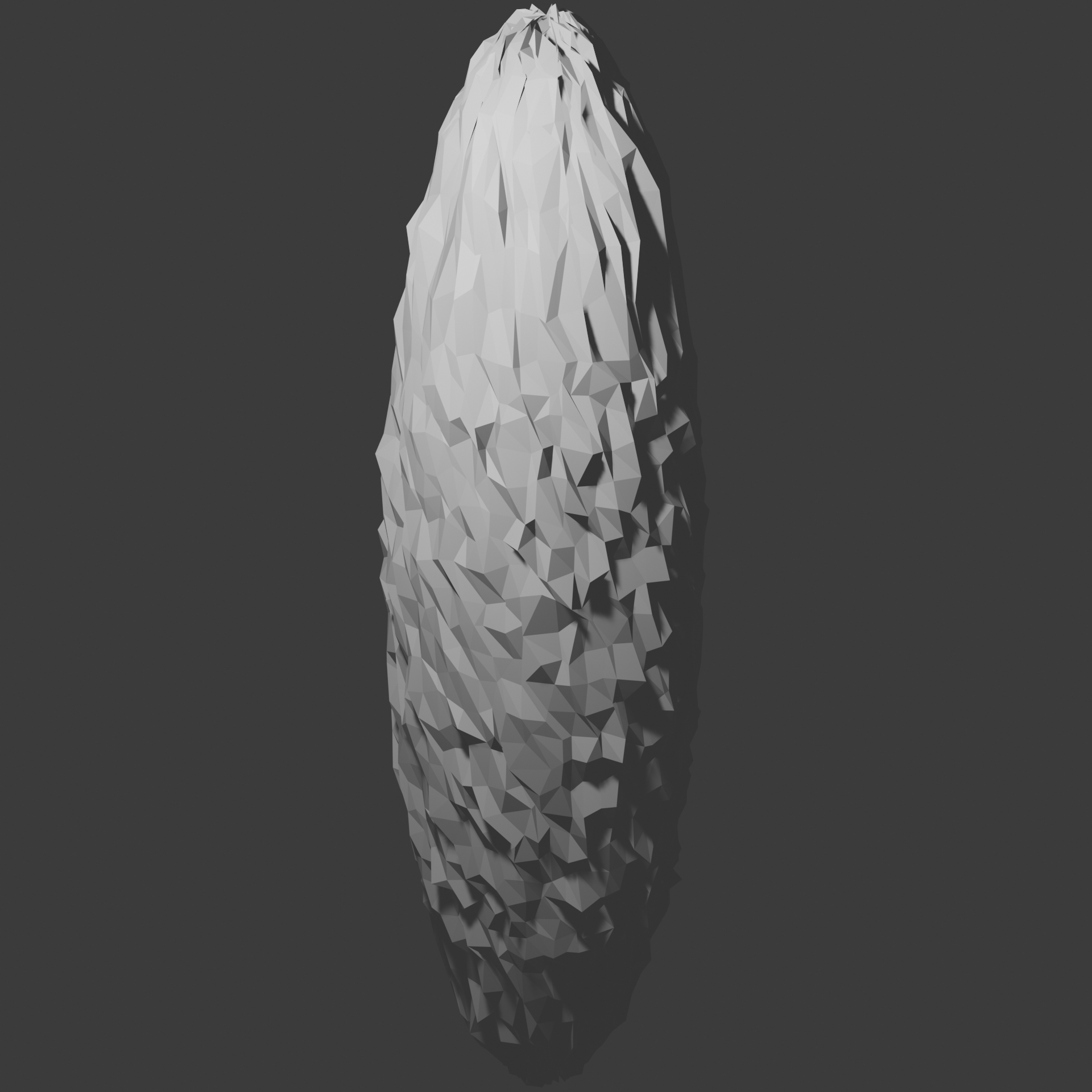}
        \label{aug:fig:b}
    \end{minipage}}
    \vspace{0.1in}
    \hspace{\tabcolsep}
    \subfloat[Radar Response]{%
        \begin{minipage}[c][1.43in]{0.22\textwidth}
        \centering
        \includegraphics[width=\textwidth]{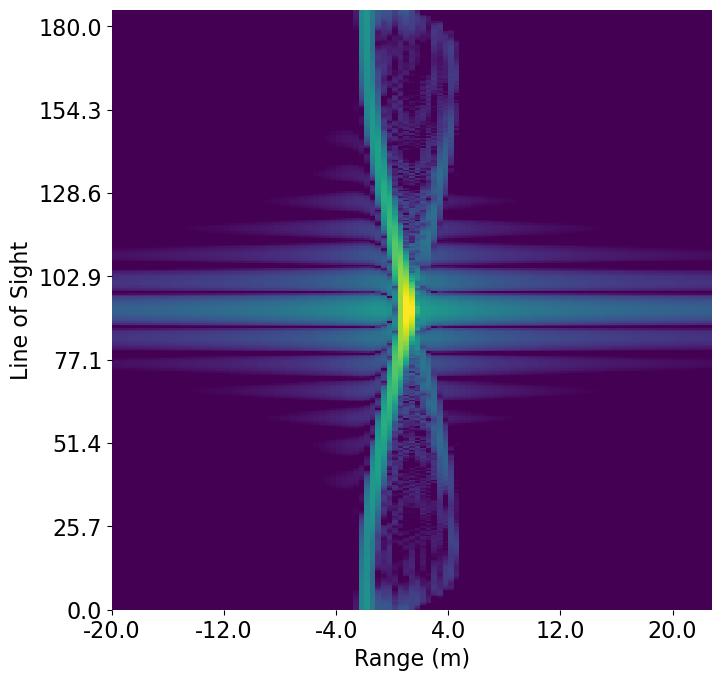}
        \label{aug:fig:c}
    \end{minipage}}
    \hspace{\tabcolsep}
    \subfloat[Augmented Response]{%
        \begin{minipage}[c][1.43in]{0.195\textwidth}
        \centering
        \includegraphics[width=\textwidth]{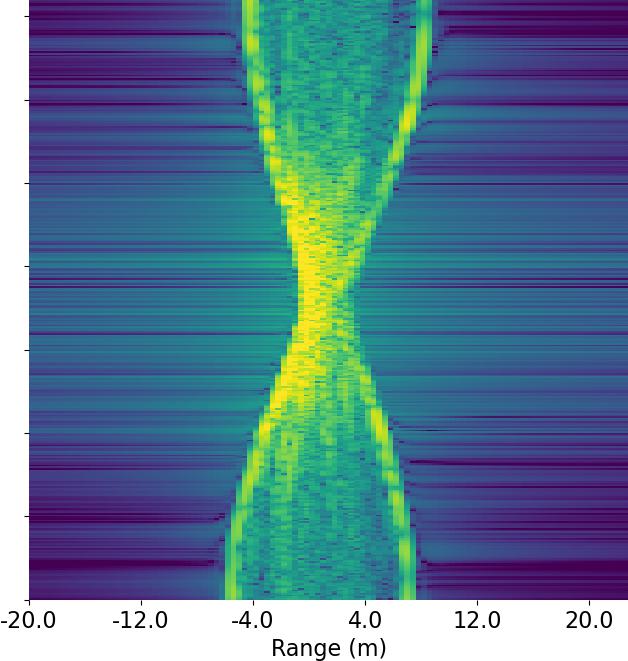}
        \label{aug:fig:d}
    \end{minipage}}
    \caption{Example radar responses generated by a first principles radar simulator. \cite{balanis2012advanced}  
    Fig. \ref{aug:fig:a} and Fig. \ref{aug:fig:c} show the mesh and radar response. Fig. \ref{aug:fig:b} shows the mesh after scaling and vertex jitter, which are both standard augmentations. Both augmentations significantly alter the radar response (Fig.\ref{aug:fig:d}).}
    \label{fig:sim}
\end{figure}

Highly accurate physics simulators operating on 3D objects and scenes exist for a variety of domains, such as optical rendering~\cite{shirley2000}, aerodynamics~\cite{hess1967}, and radar signature generation~\cite{taflove1987, andersh2000Xpatch}. Often this high accuracy is enabled by a correspondingly high computational cost~\cite{nobari2023}. This has led to a growing interest in accelerating simulations of complex physical phenomena using neural network models~\cite{kohler2023symmetric, Wheeler2017DeepSR,sanchezgonzalez2020}. 

Neural networks leverage vast amounts of data and a computationally intensive training process in order to learn a mapping between the input modality and the corresponding output. This enables subsequent simulations using fewer computational resources. The existence of large datasets containing millions of training samples and effective data augmentation strategies have allowed deep learning methods to be extremely successfully in tasks related to computer vision and natural language processing~\cite{SzegedyVISW15, devlin2018BERT}. However, generating physical simulation datasets of the same magnitude as typical deep learning datasets can be challenging.

Another challenge in training neural networks for simulation is the input data format. Physics simulators often operate on mesh representations of 3D objects~\cite{Wheeler2017, nobari2023}, which are challenging for neural network architectures to process due to their irregular structures~\cite{nobari2023}. Additionally, there can be many valid underlying mesh representations, or topologies, for a single 3D object; for example, a flat plane is typically represented with fewer triangles in a computer aided design workflow as compared to a simulation workflow. Neural networks are typically trained to be robust to variation in mesh topologies through mesh augmentation~\cite{morozov2021@mesh_aug}. However,  as demonstrated in Fig.\ \ref{fig:sim}, many physics simulators are extremely sensitive to changes in mesh shape and augmented meshes would need to be re-simulated to ensure accurate training data. Since simulation is often a computational bottleneck, this limits the applicability of common mesh augmentations to training neural network models for simulation tasks. In this work, we investigate the impact of variations in mesh topology on neural simulator performance. 
 
\begin{itemize}

\item Using radar simulation as our physics simulation task, we show that neural simulators are extremely sensitive to variations in mesh topology.

\item We compare existing methods for pretraining on large mesh datasets and find that pretraining on a mesh reconstruction objective reduces sensitivity to mesh topology.

\item We introduce Basic Shapes, a novel mesh dataset that enables direct analysis on the sensitivity of neural simulators to topology through shape-preserving variations.
\end{itemize}

\section{Related Work}

\subsection{Neural Networks for 3D Objects } 

A variety 3D object representations are used by neural networks, including meshes, point clouds, voxel occupancy, and sign distance functions \cite{feng2018meshnet,hong2024lrm}. However, since meshes are the native format for many physics simulators, there is a strong motivation to use meshes as the input for neural simulators \cite{kohler2023symmetric,Wheeler2017DeepSR,sanchezgonzalez2020}.

\begin{figure}[t!]
\centering
    \subfloat[Simple Mesh]{%
        \begin{minipage}[c][1in]{0.15\textwidth}
        \includegraphics[width=\textwidth]{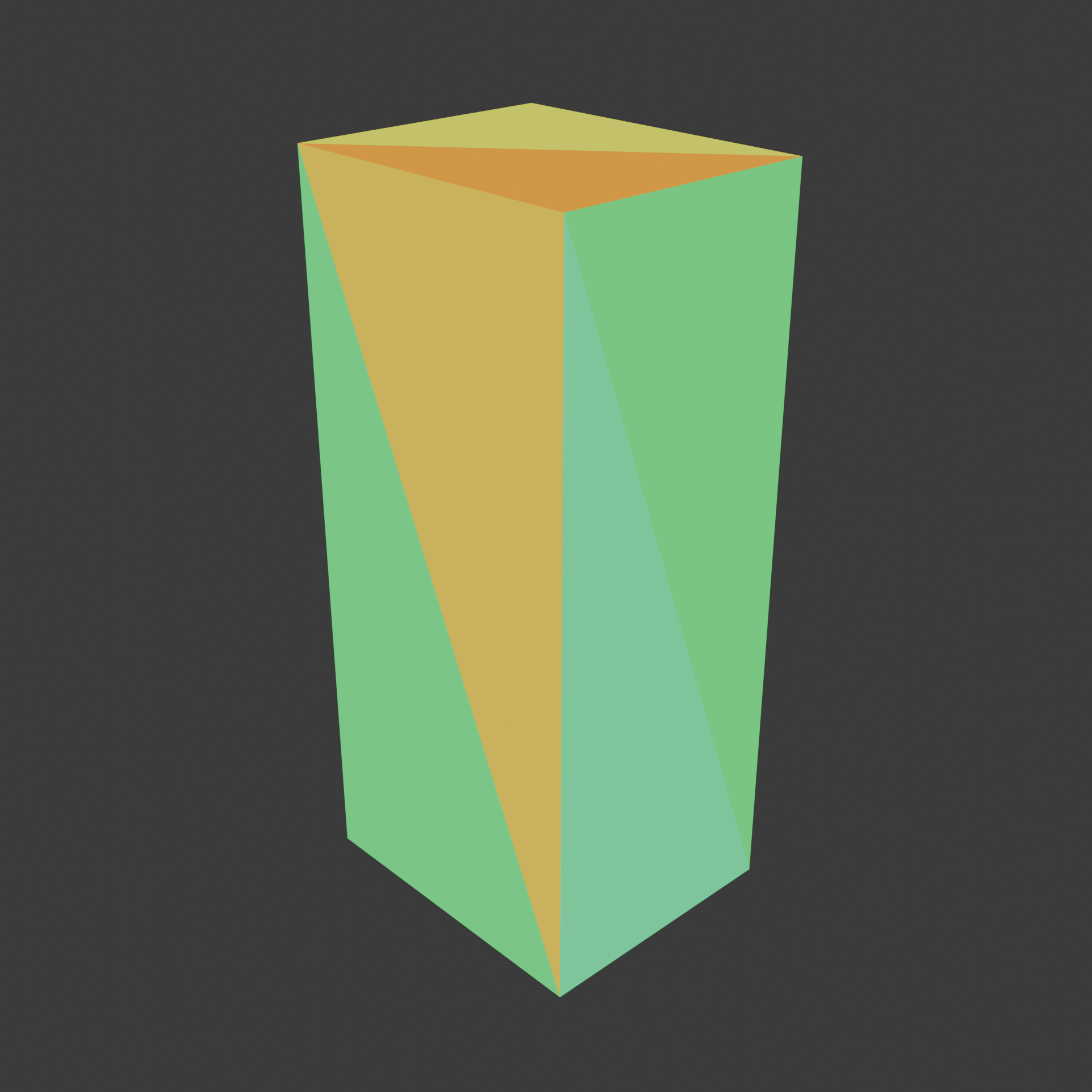}
        \label{basic:fig:a}
    \end{minipage}}
    \hspace{\tabcolsep}
    \subfloat[Complex Mesh 1]{%
        \begin{minipage}[c][1in]{0.15\textwidth}
        \includegraphics[width=\textwidth]{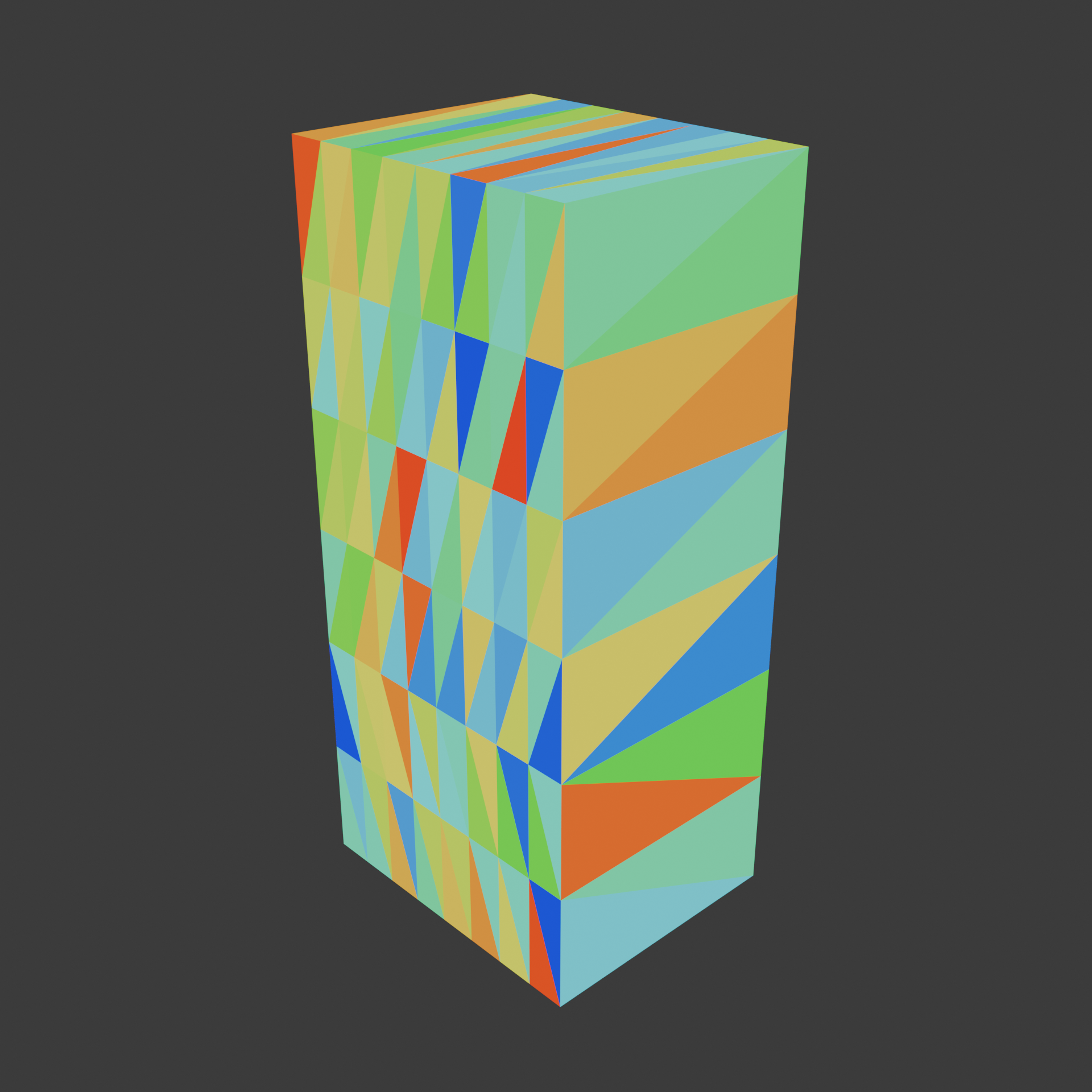}
        \label{basic:fig:b}
    \end{minipage}}
    \hspace{\tabcolsep}
    \subfloat[Complex Mesh 2]{%
        \begin{minipage}[c][1in]{0.15\textwidth}
        \includegraphics[width=\textwidth]{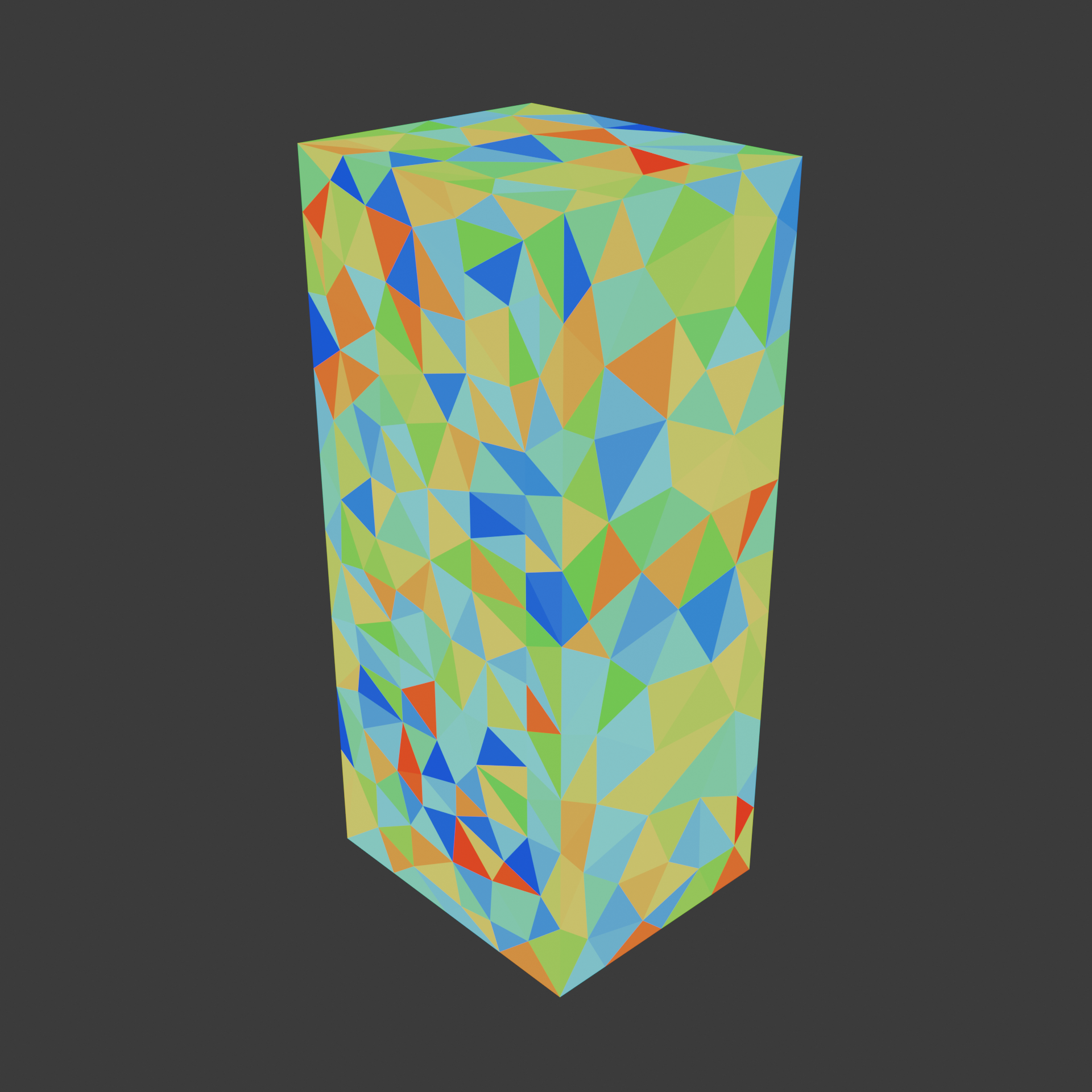}
        \label{basic:fig:c}
    \end{minipage}}
    \caption{Example object from the Basic Shapes dataset with randomly colored faces. Fig.\ \ref{basic:fig:a} shows the simple mesh representation and Fig.\ \ref{basic:fig:b} - Fig.\ \ref{basic:fig:c} show examples of shape preserving augmentations with more complex topologies.} 
    \label{fig:basic}
\end{figure}

Neural network architectures have been developed for tasks like  mesh classification, segmentation and generation ~\cite{feng2018meshnet,liang2022meshmaemaskedautoencoders3d,siddiqui2023meshgpt}. The first component in these architectures is a mesh embedding network that embeds faces as a feature vector. Direct face embedding networks calculate face features from quantities like vertex positions and face normals~\cite{hanocka_2019,siddiqui2023meshgpt}. Graph embedding networks calculate initial face features, but then utilize graph convolutions to incorporate information from nearby faces~\cite{siddiqui2023meshgpt}. Face tokenization methods pair graph embedding with a learned codebook to enable mesh reconstruction tasks~\cite{siddiqui2023meshgpt}. Once faces have been embedded, their features are aggregated to create an overall mesh feature.

There is limited work examining the impact of mesh topology on neural simulators. The closest body of work evaluates the impact of mesh augmentations on downstream tasks~\cite{morozov2021@mesh_aug} or pre-training strategies for mesh representations~\cite{siddiqui2023meshgpt, liang2022meshmaemaskedautoencoders3d}. This prior work focuses on improving mesh representations for tasks like mesh segmentation or generation, while our work examines whether existing representations are insensitive to mesh topology in the context of physics simulation tasks.

\subsection{Pretraining Neural Networks} 

 Pretraining is a technique used to improve neural network performance on a target task by first training on a related task for which more training data is available ~\cite{li2023bio}. Pretraining is effective when it is able to learn latent representations of data that generalize to the target task ~\cite{devlin2018BERT, oquab2024dinov2learningrobustvisual}
 ; due to the availability of large labeled datasets like Shapenet ~\cite{chang2015shapenet}, many styles of mesh pretraining are viable. We examine:
 \begin{itemize}
     \item \textit{Unsupervised}: Pretraining with an autoencoding mesh reconstruction task operating on local features, as in ~\cite{siddiqui2023meshgpt}.
     \item \textit{Supervised}: Pretraining to predict the semantic class of a mesh via a global classification objective ~\cite{lahav2020meshwalker, deng2009Imagenet}.
 \end{itemize}

\subsection{Radar Modeling and Simulation } 

We use radar simulation as a representative task to evaluate the impact of mesh topology on model performance. Radar is not sensitive to mesh topology but is highly sensitive to small changes in object shape, preventing the use of standard mesh augmentations in training. Radar also has applications across domains such as air traffic control and autonomous driving \cite{balanis2012advanced}, and there exists prior work on neural radar simulators \cite{Wheeler2017DeepSR, kohler2023symmetric} showing promise in neural solutions for this space.

\section{Methods}

We describe the radar simulation task, the different mesh embedding methods, and the pretraining tasks we investigated.

\subsection{Radar Simulation Task and Architecture}
We use the radar simulation task defined in \cite{kohler2023symmetric} for our experiments. Given a mesh $M$ composed of a set of vertices $V_M$ and a set of faces $F_M$, the neural simulator $g$ must output a radar response $\hat{R}$ that is similar to the ground truth simulation response $R$. The neural simulator $g$ is composed of three components; a face embedding network $f$, an aggregation network $a$, and a decoding network $d$, such that $R = d(a(f(V_M, F_M)))$, represents the radar signal output.

\subsection{Face Embedding}
The first step in generating a learned representation for meshes is typically to generate a feature embedding $e_i$ for each face $f_i \in F_M$. In our experiments, we use three different types of face embedding networks.

\subsubsection{Direct Face Embedding} 

Face embeddings are calculated from geometric face features, such as vertex positions and face normals, using standard neural network layers. We implement the transformer-based method from \cite{kohler2023symmetric} for our experiments, since it has already shown good performance on the radar simulation task.

\subsubsection{Graph-based Face Embedding} 

Direct face embeddings are refined using a graph neural network. We use the graph encoding network from \cite{siddiqui2023meshgpt}, which has been able to effectively generate features for complex mesh generation tasks \footnote{\label{note1}We utilize Phil Whang's implementation~\cite{lucidrains} of these methods.}.

\subsubsection{Tokenization-based  Embeddings} 

Codebook tokenization can be used to reduce mesh embedding size and effectively discretize mesh representations for generative tasks \cite{siddiqui2023meshgpt}; we use the codebook tokenization from \cite{siddiqui2023meshgpt}. 

\subsection{Aggregator and Decoder Networks}

Once face embeddings have been generated, a feature aggregator network $a$ maps the face information into a single feature vector  $C_M = a(\{ e_i \})$, which represents the full mesh. The decoder network $d$ then maps this to the simulated radar response $\hat{R} = d(C_M)$. The radar simulator is optimized to minimize error with respect to the true response.

For our experiments, we use a transformer for the aggregator network architecture and a MLP for the decoder \cite{transformers}. This architecture was introduced for radar simulation in~\cite{kohler2023symmetric}; however, it did not utilize scale-normalization on mesh inputs as is typical in pre-training mesh embedding models. We augment the decoder to handle scale-normalized meshes by learning a categorical embedding to represent discretized scale and concatenating this embedding with the aggregated mesh representation before decoding. We train all models using a mean squared error loss that is weighted by target intensity to emphasize important radar signature features.

\subsection{Training Methods} 

Effective mesh encodings for simulation tasks must emphasize features like object shape while ignoring irrelevant mesh topology. We use these training strategies in our experiments.

\subsubsection{Training from Scratch}  

The face embedding network $f$ is randomly initialized and trained end-to-end on the target simulation dataset to predict $\hat{R}$.

\subsubsection{Classification-based Pretraining} 

 The features generated by a mesh face embedding network are fed to a transformer encoder with a classification head \cite{transformers} that is trained to predict semantic class labels on a larger auxiliary mesh data set. Since radar response is not needed, standard mesh augmentation techniques can be applied during training. The weights of the embedding network are used to initialize $f$, which is then trained end-to-end to predict $\hat{R}$.

\subsubsection{Autoencoder-based Pretraining} 

Mesh face embedding networks are trained in the augmented auxiliary dataset, using an objective of the autoencoder based on mesh reconstruction \cite{siddiqui2023meshgpt}. The weights of the embedding network are used to initialize $f$, which is then trained end-to-end to predict $\hat{R}$. 
 
\subsection{Data Generation}

To evaluate the impact of topology variations on downstream simulation tasks, we need a set of objects paired with various mesh instantiations that map to the same simulation output. Without this property, it is difficult to isolate the impact of mesh variation from the general simulation capability. There are many 3D object datasets that contain wide varieties of meshes such as Shapenet~\cite{chang2015shapenet} and Objaverse~\cite{deitke2022objaverse}; however, these do not provide multiple mesh topologies for individual objects. 

To address this gap, we propose a new dataset \textbf{Basic Shapes.}  We use Blender~\cite{blender}, an open source 3D modeling and rendering program, along with the following process to generate simple and complex mesh topologies with exactly the same underlying shape. First, we select a primitive 3D shape from Blender's built-in defaults, generate a mesh to represent the object, and scale it by random values along the independent object axes. We ensure that no primitive of the same type is generated with a similar scale to ensure object diversity. This mesh is considered the \textbf{simple} mesh for that object. Next, we randomly add loop cuts to create additional faces and follow up with a decimation operation to reduce faces in a pseudorandom manner~\cite{blender}. We carefully choose the decimation level so that the shape of the underlying object is not changed. We also randomly vary the number of vertices used to define the curvature for primitives that have curved components. We repeat this process by varying the number of loop cuts, decimation parameters, and curvature vertices until the desired number of \textbf{complex} mesh variations has been generated; examples can be seen in Fig. \ref{fig:basic}. 

Using this process, we generate the \textbf{Basic Shapes} dataset to enable analysis on the impact of variations in mesh topologies that preserve object shape. This dataset consists of Cube, Cylinder, and Sphere primitives, where each class contains 1000 different objects at different scales. For each object, a single \textbf{simple} mesh and 99 \textbf{complex} mesh variants are generated, for a total of 100 mesh instantiations. In total, we generate 3000 objects and 300,000 meshes for this dataset,  split 90/10 between training and testing. 

\section{Experiments}

\subsection{Datasets}

We follow the procedure in \cite{kohler2023symmetric} for simulating radar responses, but use objects from the Basic Shapes dataset as our mesh source. We then separate the simple mesh topologies from the complex mesh variants; the simple meshes are used for training models, while complex mesh variants are used only for evaluation and an ideal training scenario. For pretraining, we follow the process described in~\cite{siddiqui2023meshgpt} to prepare Shapenet, resulting in 20,789 decimated Shapenet objects~\cite{chang2015shapenet}.

\subsection{Metrics}

We use three variants of $MSE$ \footnote{MSE (Mean Square Error): $= \frac{1}{n} \sum_{i=1}^{n} (R_i - \hat{R}_i)^2$.} to evaluate the impact of mesh artifacts on the quality of a predicted radar response \cite{kohler2023symmetric}. 
\begin{align*}
\textrm{\textbf{Simple MSE}} &: MSE(R, \hat{R}_s), \\
\textrm{\textbf{Complex MSE}} &: \frac{1}{|C|}\sum_{c\in C} MSE(R, \hat{R}_c), \\
\textrm{\textbf{Variation MSE}} &: \frac{1}{|C|}\sum_{c\in C} MSE(\hat{R}_s, \hat{R}_c).
\end{align*}
where $\hat{R}_s$ is the neural network prediction for the simple mesh $s$ and $\hat{R}_c$ is a  prediction for a complex mesh in the set of complex mesh variants $C$. Simple MSE and Complex MSE measure simulation accuracy for simple and complex topologies. Variation MSE directly measures the impact of mesh topology by comparing the predicted responses for simple and complex meshes that represent the same shape. A low MSE value on Complex and Variation MSE indicates that mesh topology has less impact neural network predictions. 

\subsection{Training and Model information}

Models were pretrained on Shapenet using 30 V100 gpus for 50 epoches ($\sim$1 day). Simulation models for each established combination of embedding method and training condition were trained for 50 epochs on the simple meshes in Basic Shapes \footnote{Data and code available here: \href{https://github.com/nathan-vaska/mesh_topology}{\textcolor{blue}{github.com/nathan-vaska/mesh\_topology}}} using two V100 gpus ($\sim$3 hours). Models were also trained from scratch on all meshes to provided an idealized comparison; these models were trained on 2 V100 gpus for 5 epoches ($\sim$1 day).

\section{Results and Discussion}

\begin{figure}[t!]
\centering
\vspace{-0.15in}
\hspace{0.15in}
    \subfloat[Object]{%
        \begin{minipage}[c][1.2in]{0.14\textwidth}
        \centering
        \includegraphics[width=\textwidth]{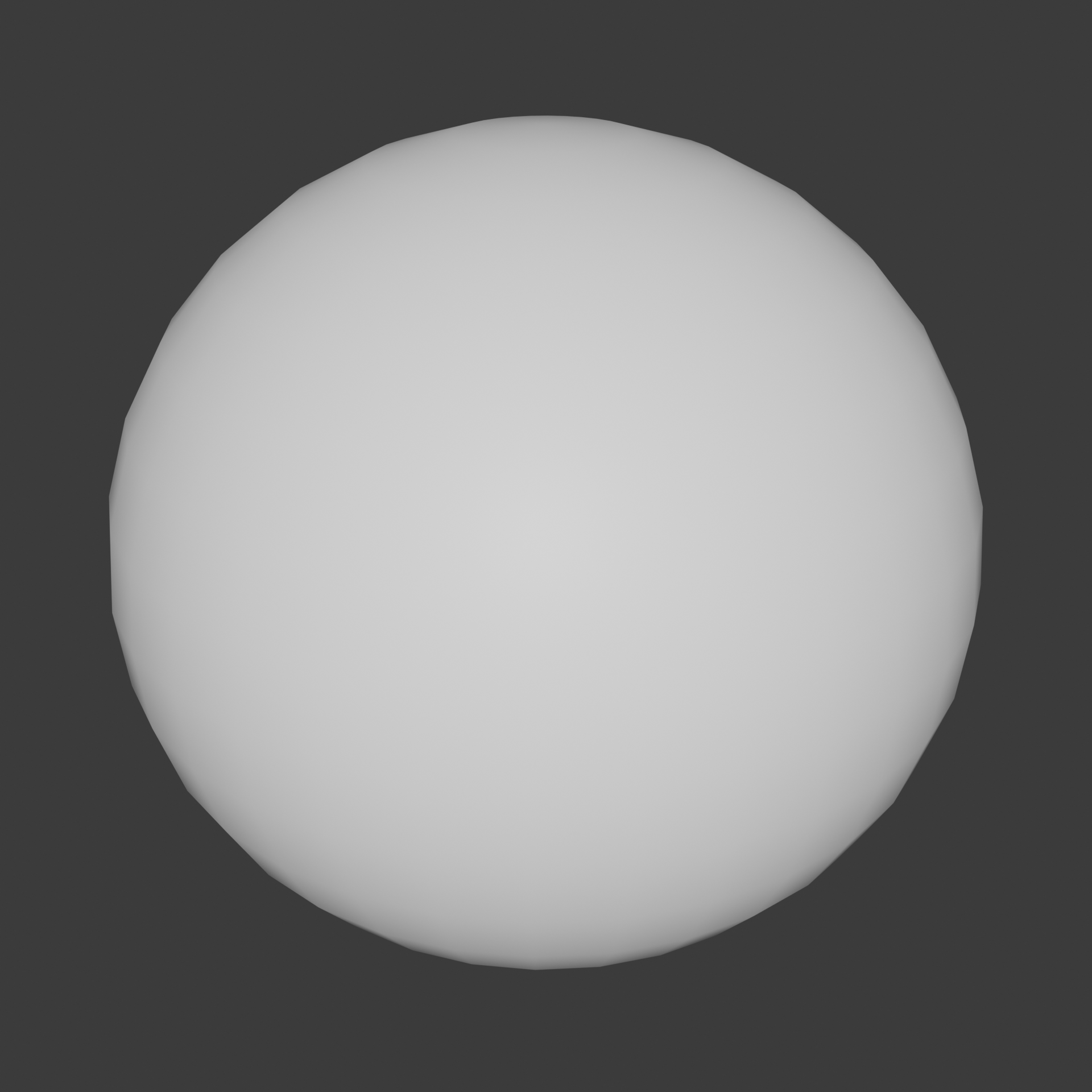}
    \end{minipage}}
    \hspace{\tabcolsep}
    \subfloat[Simple Mesh]{%
    \begin{minipage}[c][1.2in]{0.14\textwidth}
        \centering
        \includegraphics[width=\textwidth]{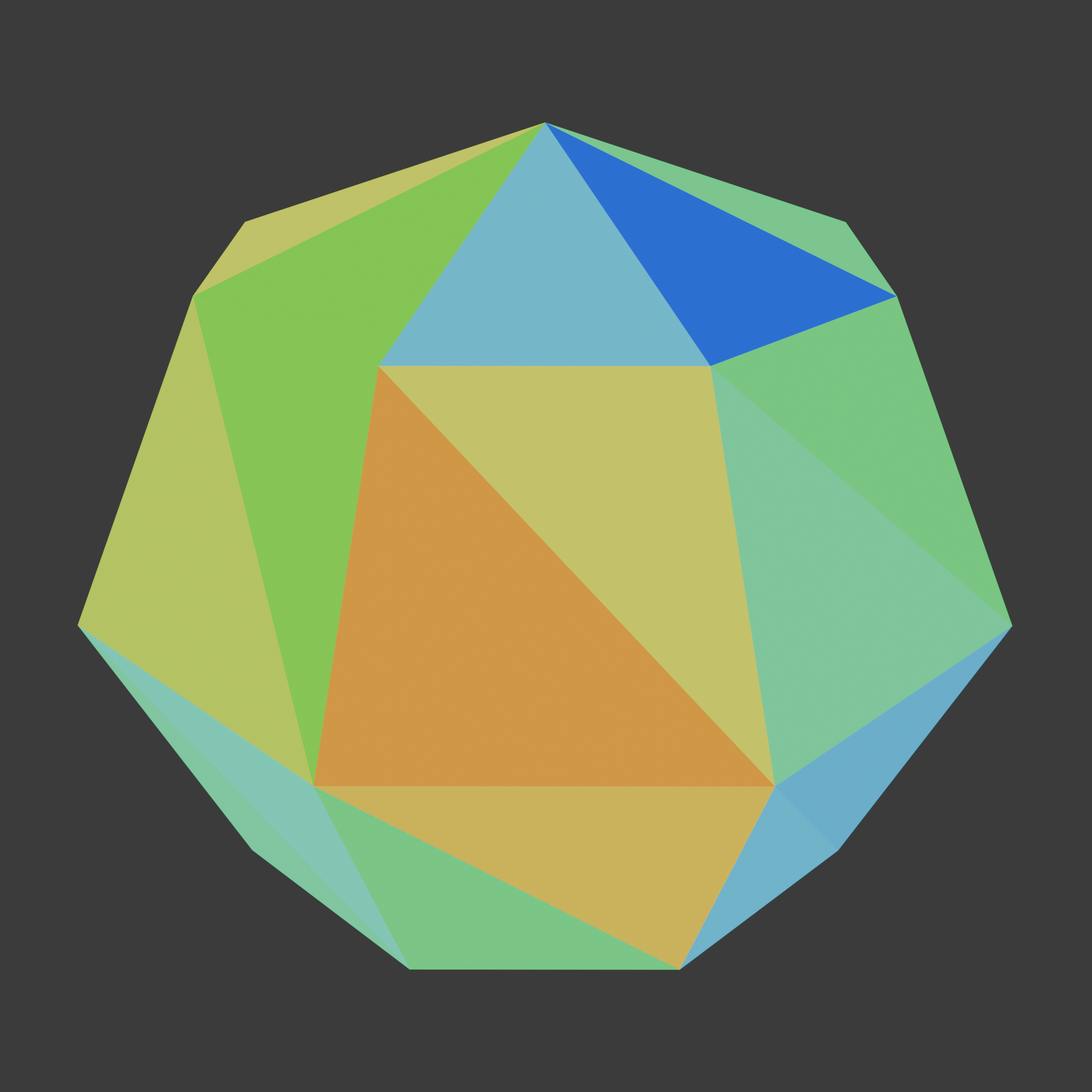}
    \end{minipage}}%
    \hspace{\tabcolsep}
    \subfloat[Complex Mesh]{%
       \begin{minipage}[c][1.2in]{0.14\textwidth}
        \centering
        \includegraphics[width=\textwidth]{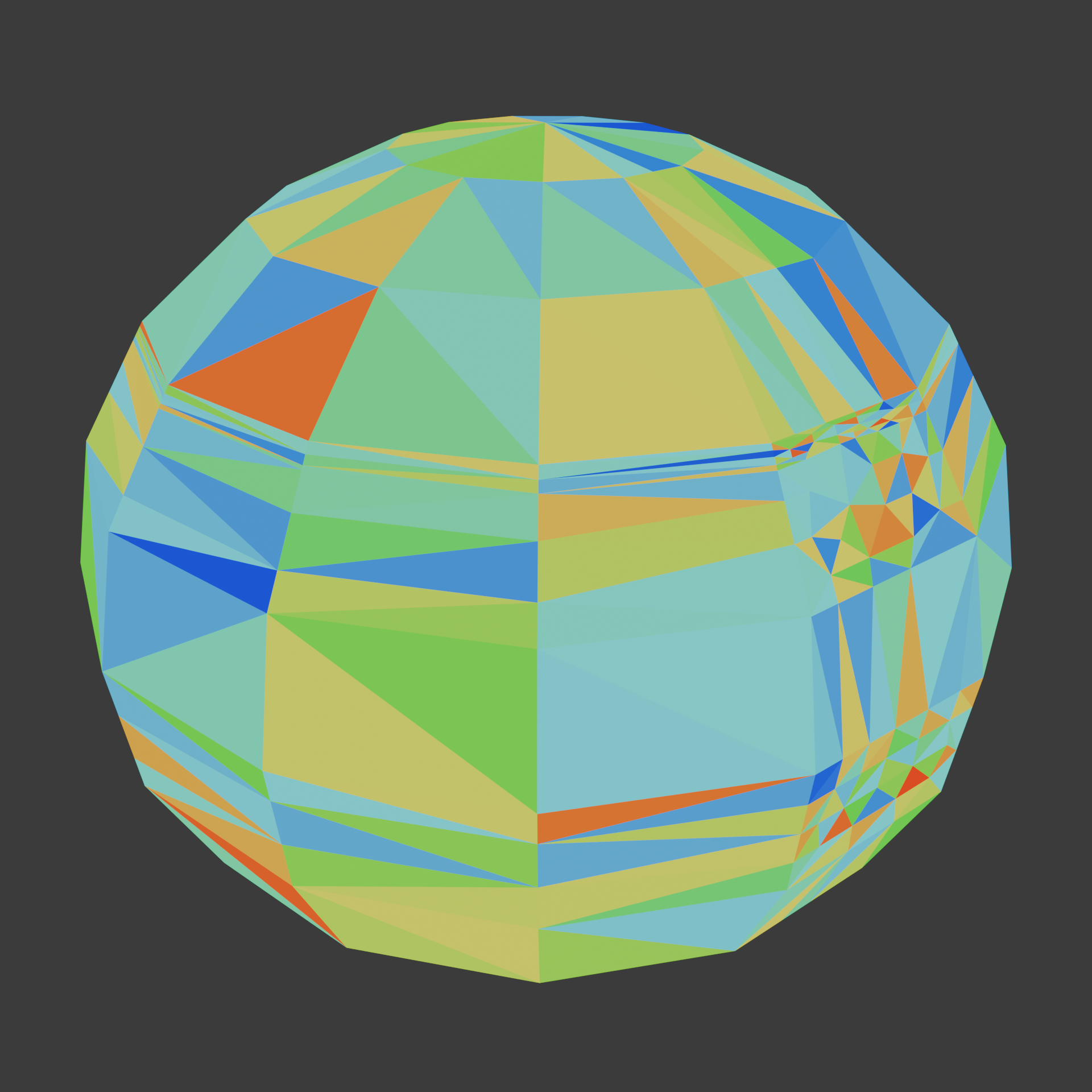}
    \end{minipage}}%
    \vspace{0.1in}
    
    \subfloat[Physics Sim.]{%
        \begin{minipage}[c][1.2in]{0.164\textwidth}
        \centering
        \includegraphics[width=\textwidth]{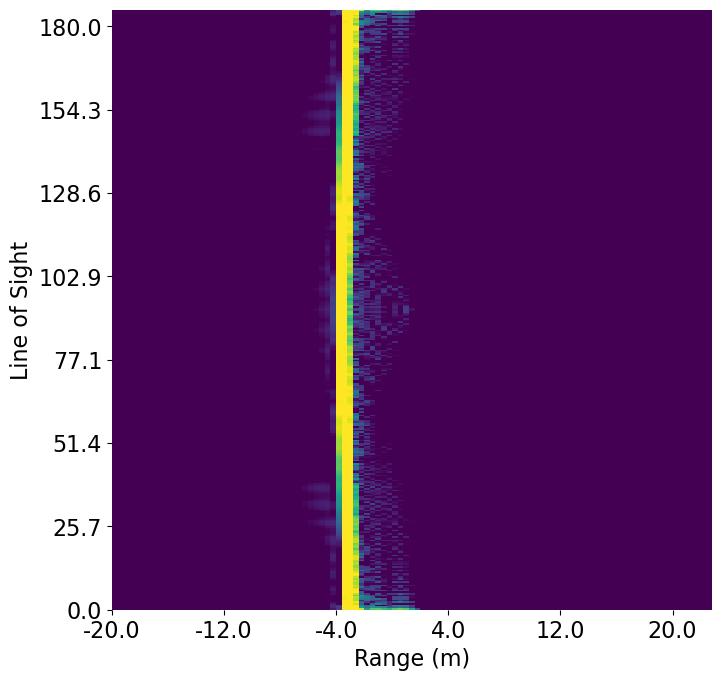}
    \end{minipage}}
    \hspace{\tabcolsep}
    \subfloat[Simple Pred.]{%
        \begin{minipage}[c][1.2in]{0.1425\textwidth}
        \centering
        \includegraphics[width=\textwidth]{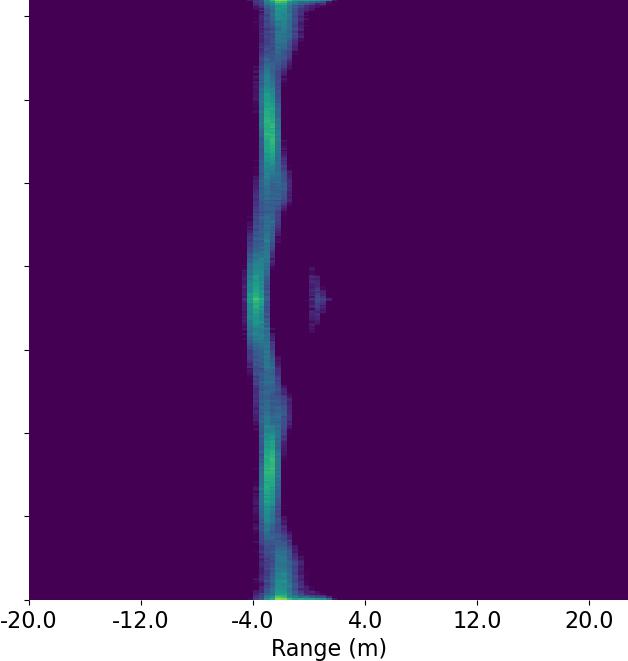}
    \end{minipage}}
    \hspace{\tabcolsep}
    \subfloat[Complex Pred.]{%
        \begin{minipage}[c][1.2in]{0.1425\textwidth}
        \centering
        \includegraphics[width=\textwidth]{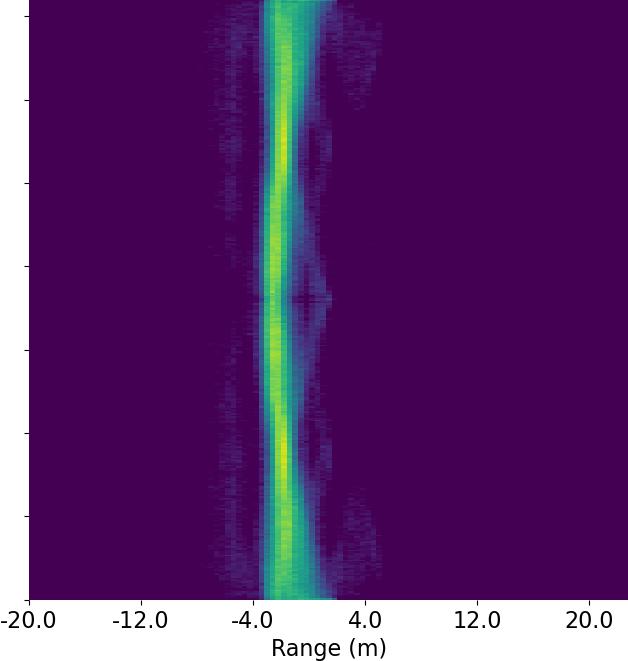}
    \end{minipage}}
    \caption{Example predictions from graph embedding model pretrained via the auto-encoding objective. The model accurately predicts the targets response for simple meshes, and predicts a similar response for complex mesh variants despite significantly increased complexity.}
\end{figure}

Table~\ref{tab:res} shows the performance of each algorithm on each metric. We observe that the graph embedding model pre-trained with the autoencoding objective had the most accurate predictions on both simple and complex meshes, while also performing best on the variation metric. In general, algorithms using graph or tokenization embedding performed better than algorithms using direct embedding, and autoencoder training also generally outperformed classification pretraining or no pre-training. Taken together, these results indicate that graph-based encoders and autoencoder pre-training are more effective at handling complex and varied mesh topology. There is potentially a theoretical underpinnings to these results; both graph embeddings and the autoencoding pretraining objective are more strongly tied to local spatial features, which are extremely important for accurate simulation in physics- based radar simulators \cite{taflove1987}, \cite{andersh2000Xpatch}. Additionally, the pre-training exposes algorithms to additional meshes and their topologies, likely improving their feature quality on complex mesh topologies. Since local features are important to a variety of simulators in other domains \cite{shirley2000}, \cite{hess1967}, we expect that better mesh embeddings for the radar simulation task will transfer well to such domains.

While the autoencoding-pretrained graph embedding architecture is the least sensitive to mesh topologies of the methods tested, it is still significantly more sensitive than models trained on the idealized dataset that includes typically unavailable topology augmentations. This indicates that the sensitivity of these architectures to mesh topology can be further reduced. In particular, the pre-training objectives might prove to be more effective when trained against a larger mesh dataset (the dataset size considered here is significantly lower to other domains, like computer vision and NLP, where significant improvements due to pre-training have been observed \cite{devlin2018BERT} \cite{oquab2024dinov2learningrobustvisual}), altering the pre-training loss function to enforce similarity between similar meshes, or by incorporating equivariance to mesh orientations into the mesh embedding structures \cite{kohler2023symmetric}. 

\begin{table}[t!]
\begin{center}
\resizebox{\columnwidth}{!}{%
\begin{tabular}{|c|c|c||c|c|c|}
\hline
\multicolumn{3}{|c||}{\textbf{Model Details}}&\multicolumn{3}{|c|}{\textbf{Metrics}} \\
\hline
\textbf{Mesh}&\textbf{Pretraining}&{\textbf{Training}}&\textbf{Simple}&\textbf{Complex}&{\textbf{Variation}} \\
\textbf{Encoder} & \textbf{Style}& \textbf{Data}& \textbf{MSE}& \textbf{MSE}& \textbf{MSE} \\
\hline
Direct & None      & Ideal   & 66.43& 67.8& 2.716\\
Graph  & None      & Ideal   & 66.9& 66.8& 1.2\\
Token  & None      & Ideal   & 90.5& 90.5& 5.8\\
\hline
\hline
Direct & None & Basic    & 159.3 & 192.5 &  \textasteriskcentered \\ 
Graph  & None & Basic    & 61.2  & 175.4 & 216.6  \\
Token  & None & Basic    & 75.4  & 200.4 & 321.3  \\
\hline 
Direct & Classification  & Shapenet & 217.2 & 217.2 & \textasteriskcentered \\ 
Graph  & Classification  & Shapenet & 180.3 & 233.9 & 289.7  \\
Token  & Classification  & Shapenet & 75.61 & 215.2 & 178.2 \\
\hline
Graph  & Autoencoder & Shapenet & \textbf{58.2} & \textbf{164.1} & \textbf{125.6}  \\
Token  & Autoencoder & Shapenet & 65.3 & 176.3 & 143.1 \\
\hline
\end{tabular}
}
\end{center}
\caption{Performance of models on the radar simulation task; \textasteriskcentered \  indicates a low Variation MSE induced by mode collapse to a single radar response, rather than accurate predictions.}

\label{tab:res}
\end{table}

\section{Conclusion}
Many high-fidelity physics simulators use mesh representations as inputs, and there is a growing interest in using neural networks to reduce their high computational cost. In parallel, there have been advancements in neural network models which work directly on mesh data. However, the unique requirements of physics simulation as compared to other mesh-based tasks limit the use of common mesh augmentations. Utilizing the new Basic Shapes dataset that we introduce in this work, we show that the performance of neural simulators trained without these augmentations is degraded by irrelevant variations in mesh topology. We find that pre-training graph-based mesh embedding networks on an autoencoding task effectively reduces this sensitivity to mesh topology, and suggest additional avenues of research to continue reducing this sensitivity. 

\bibliographystyle{IEEEtran}
\bibliography{references}

\end{document}